\documentclass[conference]{IEEEtran}
\IEEEoverridecommandlockouts
\usepackage{cite}
\usepackage{amsmath,amssymb,amsfonts}
\usepackage{algorithmic}
\usepackage{graphicx}
\usepackage{textcomp}
\usepackage{url}
\usepackage{cite}
\usepackage{booktabs}

\usepackage{fancyhdr}

\fancypagestyle{firstpage}{
    \fancyhf{}
    \fancyfoot[C]{\footnotesize This paper has been accepted for presentation at the IEEE International Conference on Acoustics, Speech, and Signal Processing (ICASSP) 2025. ©2025 IEEE. Personal use of this material is permitted. Permission from IEEE must be obtained for all other uses.}
}

\usepackage{xcolor}
\def\BibTeX{{\rm B\kern-.05em{\sc i\kern-.025em b}\kern-.08em
    T\kern-.1667em\lower.7ex\hbox{E}\kern-.125emX}}
\begin{document}

\makeatletter
\newcommand{\newlineauthors}{%
  \end{@IEEEauthorhalign}\hfill\mbox{}\par
  \mbox{}\hfill\begin{@IEEEauthorhalign}
}
\makeatother

\title{Audio Array-Based 3D UAV Trajectory Estimation with LiDAR Pseudo-Labeling
\thanks{This work is supported by National Research Foundation, Singapore, under
its Medium-Sized Center for Advanced Robotics Technology Innovation.}
}

\author{
\IEEEauthorblockN{Allen H.-X. Lei}
\IEEEauthorblockA{
\textit{Nanyang Technological University,}\\
Singapore. \\
allenlei0827@gmail.com}
\and
\IEEEauthorblockN{Tianchen Deng}
\IEEEauthorblockA{
\textit{Nanyang Technological University,}\\
Singapore. \\
n2308684a@e.ntu.edu.sg}
\and
\IEEEauthorblockN{Han Wang}
\IEEEauthorblockA{
\textit{Nanyang Technological University,}\\
Singapore. \\
wh200720041@gmail.com}
\newlineauthors
\IEEEauthorblockN{Jianfei Yang}
\IEEEauthorblockA{
\textit{Nanyang Technological University, Harvard University,}\\
Singapore, USA. \\
jianfei.yang@ntu.edu.sg}
\and
\IEEEauthorblockN{Shenghai Yuan*}
\IEEEauthorblockA{
\textit{Nanyang Technological University,}\\
Singapore. \\
*Corresponding author: shyuan@ntu.edu.sg}
}

\maketitle

\begin{abstract}
As small unmanned aerial vehicles (UAVs) become increasingly prevalent, there is growing concern regarding their impact on public safety and privacy, highlighting the need for advanced tracking and trajectory estimation solutions. In response, this paper introduces a novel framework that utilizes audio array for 3D UAV trajectory estimation. Our approach incorporates a self-supervised learning model, starting with the conversion of audio data into mel-spectrograms, which are analyzed through an encoder to extract crucial temporal and spectral information. Simultaneously, UAV trajectories are estimated using LiDAR point clouds via unsupervised methods. These LiDAR-based estimations act as pseudo labels, enabling the training of an Audio Perception Network without requiring labeled data. In this architecture, the LiDAR-based system operates as the Teacher Network, guiding the Audio Perception Network, which serves as the Student Network. Once trained, the model can independently predict 3D trajectories using only audio signals, with no need for LiDAR data or external ground truth during deployment. To further enhance precision, we apply Gaussian Process modeling for improved spatiotemporal tracking. Our method delivers top-tier performance on the MMAUD dataset, establishing a new benchmark in trajectory estimation using self-supervised learning techniques without reliance on ground truth annotations.
\end{abstract}


\section{Introduction} \label{sec:intro}

 In recent years, Unmanned Aerial Vehicles (UAV) \cite{yuan2021survey,er2013development,ji2024lio} have become widely accessible \cite{cao2020online,cao2025cooperative}, raising significant security concerns such as privacy violations \cite{lyu2022structure, wang2017heterogeneous}, unauthorized surveillance \cite{cao2022direct, liu2024distance,lyu2023spins,lyu2021vision}, and risks in restricted airspace \cite{cao2023doublebee, esfahani2019towards, cao2023path}, including threats to civilian aviation. Their use in cross-border drug smuggling \cite{esfahani2020unsupervised} and military operations \cite{wu2020achieving,xu2024m} underscores the urgent need for effective UAV tracking technologies \cite{yuan2014Autonomous}.

Despite the pressing need for UAV tracking systems, existing studies predominantly rely on visual information \cite{seidaliyeva2020real, yang2022overcoming,zheng2021air,isaac2021unmanned,liu2021real,lai2024nvp,jin2024robust}   for 2D tracking or 3D detection from RADAR/LiDAR \cite{vrba2023onboard,liang2024separatingdronepointclouds} \cite{deng2024multimodaluavdetectionclassification,liang2024unsuperviseduav3dtrajectories,wang2017heterogeneous,liu2023non,yin2023segregator,nguyen2024uloc,10530350,yuan2024large,li2024jacquard}. However, these approaches are limited by poor visibility, restricted field of view, and the high cost of RADAR/LiDAR for large-area surveillance and low-altitude smuggling UAV.\\
 \indent The utilization of audio data, despite its accessibility and cost-effectiveness, remains largely unexplored \cite{xiao2024avdtec, 10023792,10299579} in UAV trajectory estimation\cite{xiao2024tametemporalaudiobasedmamba,yang2023av}. Given the affordability and ubiquity of microphones compared to LiDAR and cameras \cite{nguyen2023vr,chen2024ig,10806842,li2024ua}, there is a significant opportunity to develop innovative 3D tracking solutions \cite{esfahani2021learning,esfahani2018new,esfahani2020local,ji2022robust,chen2024salient} that leverage audio information.

The main challenges in audio-based UAV 3D trajectory estimation are data annotation and filtering out ambient noise, which is particularly difficult in real-world scenarios where annotated data is not readily available. To address this, we propose an Audio Array-Based 3D UAV Trajectory Estimation with LiDAR to generate unsupervised pseudo-ground truth labels for self-supervised learning. During deployment, only audio is used, with Gaussian Process smoothing applied to ensure continuous trajectory estimation.  Our contributions are summarized as follows:

%


\begin{figure}[t]
\centering
\includegraphics[width=0.4\textwidth]{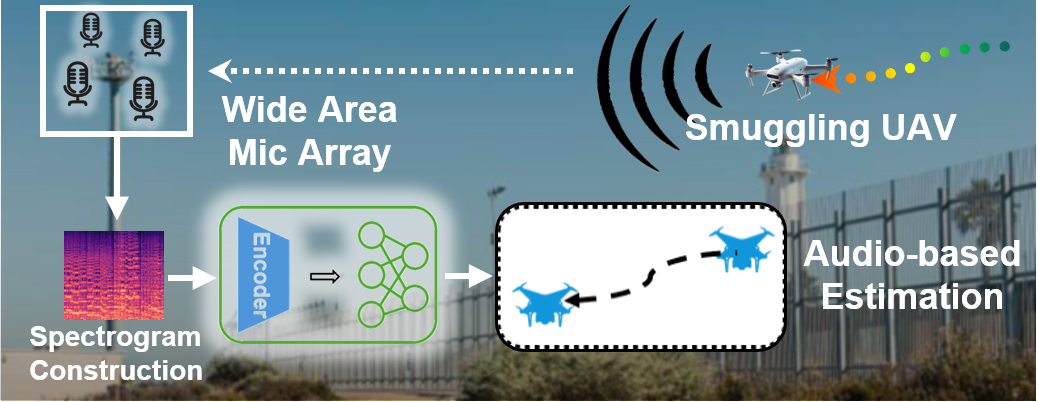}
\vspace{-1.0em}
\caption{System Motivation and Overview.}
\label{fig:introduction}
\vspace{-2em}
\end{figure}

\begin{figure*}[t]
\centering
\includegraphics[width=\textwidth]{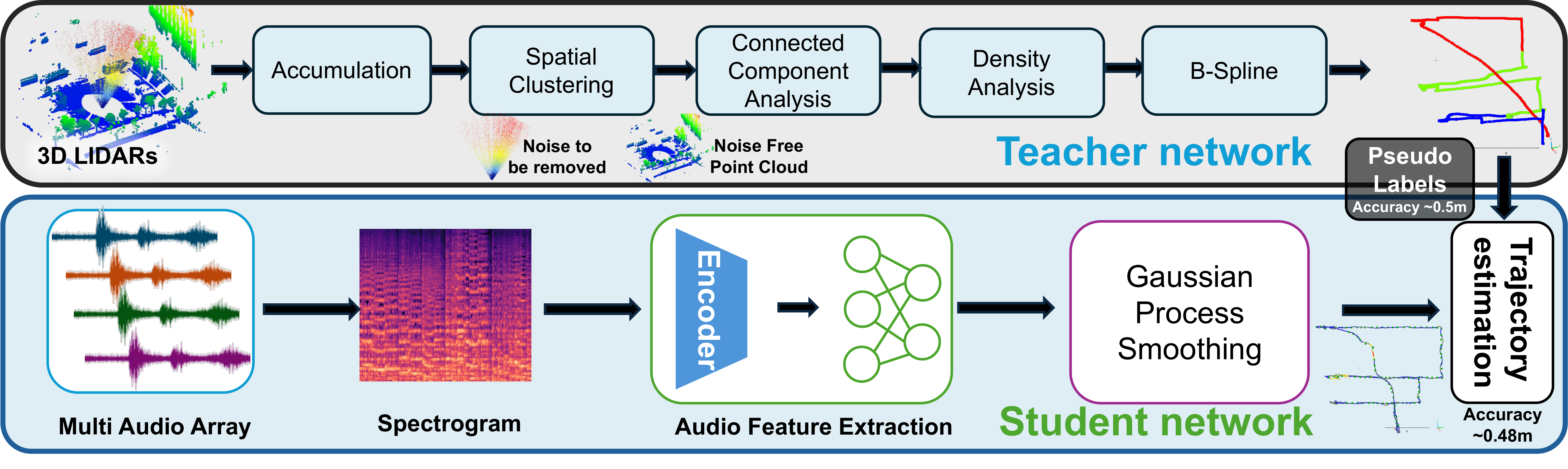}
\vspace{-2.5em}
\caption{Proposed Self-Supervised Framework. }
\label{fig:network}
\vspace{-1.7em}
\end{figure*}
\vspace{-0.25em}

\begin{enumerate}
\item We present a novel self-supervised Teacher-Student Network for audio vectoring-based 3D trajectory estimation, enabling robust UAV tracking across diverse lighting conditions without manual labels.

\item We propose a cost-effective unsupervised method for 3D drone trajectory estimation using LiDAR scans as pseudo-ground truth for the Audio Perception Network.

\item We benchmarked our method against various SOTA approaches, achieving the highest accuracy in inferring drone 3D trajectories.

\item We are open-sourcing our work to benefit the community. \url{https://github.com/AllenLei666/AAUTE}.
\end{enumerate}

\vspace{-0.25em}
\section{Proposed Method}
\vspace{-5pt}
 We introduce a novel framework for Unmanned Aerial Vehicle (UAV) trajectory estimation, which is combined with a Teacher Network and a Student Network, as illustrated in Fig. \ref{fig:network}. For the Student Network, the input consists of audio waves collected by the microphone vector. The microphone vector captures audio information pertaining to the UAV. The process begins with converting the audio waves into audio spectrograms to represent the temporal and spectral characteristics of the sound. Next, an encoder is utilized to extract features from the audio spectrograms, capturing both temporal and frequency-domain information. Then, the Gaussian Process Smoothing is implemented to smooth the estimated trajectory.

For the Teacher Network, two types of LiDAR data are processed and fused using an unsupervised learning method to estimate the 3D trajectory of UAVs. This high-precision trajectory estimate serves as a reference for the subsequent self-supervised learning phase. The LiDAR-estimated trajectory serves as pseudo labels to train the framework, enabling accurate UAV trajectory prediction from audio alone, with results cross-checked against ground truth data.

\vspace{-5pt}
\subsection{Unsupervised Teacher Network}
\vspace{-5pt}
 The Livox Mid360 provides a 360° horizontal field of view and a 59° vertical field of view above the ground, capturing both background and UAV point clouds up to 40 m. Since the LiDAR data contains significant noise, our main goal is to remove the noise and extract the UAV's LiDAR data. Initially, we need to isolate the UAV point clouds from the background. Based on recent approaches\cite{deng2024multimodaluavdetectionclassification}, we use 20 consecutive LiDAR scans as a chunk and apply the DBSCAN clustering algorithm with eps = 1.0 and min-samples = 10 to group the points into clusters for best performance. The connected component analysis is then implemented to merge adjacent clusters with distance-threshold = 2.5 and min-component-size = 20 for best-extracting performance to form larger coherent regions. The largest and smallest clusters are removed as they are typically considered noise and partial background, respectively. To further refine the UAV point cloud, we analyze the density of each cluster and select the cluster corresponding to the UAV's point cloud. 

The DJI Livox Avia LiDAR, with its 70-degree conical upward-facing field of view, tracks UAV movement up to 350m. To process the Avia data, we use the farthest point sampling technique to reduce the number of points while preserving the spatial distribution of the UAV point cloud. The dynamic part of the Livox Mid360 point cloud is then extracted and fused with the downsampled Avia data to form the complete UAV point cloud. DBSCAN is applied again to remove any remaining solar noise. Finally, with the complete UAV point cloud, B-Spline fitting is utilized to interpolate a smooth curve through the 3D UAV point cloud, generating pseudo labels that represent the UAV's trajectory with the mean error of 0.5 m compared to the real UAV's trajectory.
\vspace{-5pt}
\subsection{Self-Supervised Audio Feature Learning}
\vspace{-5pt}
 To process the audio information, we begin by collecting audio waves from a vector of microphone. The audio signals are then segmented into 2-second clips and sampled at a rate of 48,000 samples per second. Subsequently, the audio segments are converted into mel-spectrograms using a sweep window size of 2048 and a hop length of 1024. Finally, the mel-spectrograms are resized to 64 \(\times\) 64 for efficient feature extraction and integration into the training process.

The main idea behind using audio information to locate UAVs is the Time Difference of Arrival (TDOA). Each mel-spectrogram contains information from four microphones, and we primarily utilize the TDOA among these microphones to determine the UAV location. To this end, one type of convolutional kernel sweeps the spectrogram horizontally to extract information about time differences, which reflects the relative arrival times of the UAV sound at different microphones. Another type of convolutional kernel scans the spectrogram vertically to capture intensity information across different frequencies within the same spectrogram. The time-domain feature and the frequency-domain feature are then concatenated to form the audio feature. After that, we add two more fully connected layers to learn the features and change the dimension of the concatenated feature \(F_{AV}\) to 3 for trajectory estimation.


\begin{figure}[t]
\centering
\includegraphics[width=0.45\textwidth]{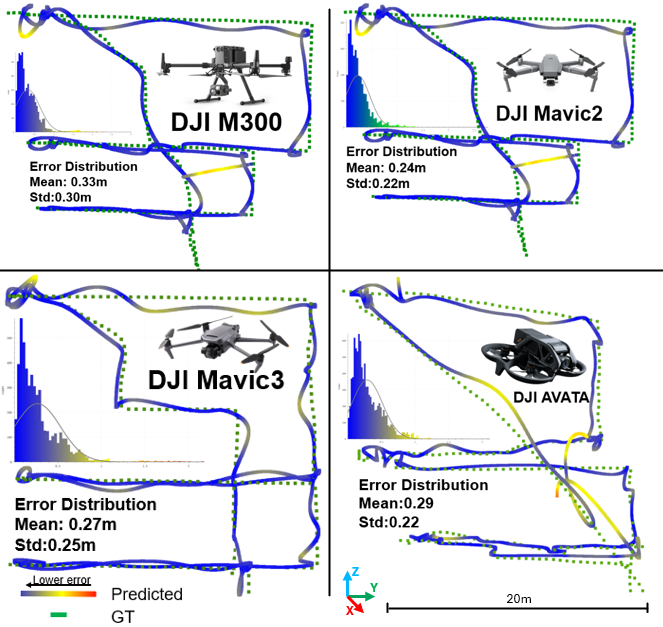}
\vspace{-1em}
\caption{Estimated Trajectory by our solution}
\label{fig:result}
\vspace{-2em}
\end{figure}

\vspace{-5pt}
\subsection{Gaussian Process Smoothing}
\vspace{-5pt}
 To enhance the smoothness of the UAV trajectory, we use Gaussian Processe (GPs) \cite{johnson2024continuoustimetrajectoryestimationcomparative}, which effectively capture complex relationships and uncertainties without relying on fixed knot lengths like B-splines.

Given a set of estimated trajectory points \( \{(t_i, \mathbf{y}_i)\}_{i=1}^N \), where \( t_i \) represents the time stamp and \( \mathbf{y}_i \) is the corresponding 3D position of the UAV at time \( t_i \), the goal is to use GP to predict the smooth trajectory.

The Gaussian Process regression model is trained using the trajectory points \( \mathbf{y}_i \) as the target variable and the timestamps \( t_i \) as the input variable. The predictive distribution on a new timestamp \( t_* \) is given by:
\vspace{-5pt}
\[
\mathbf{f}_* \mid \mathbf{t}, \mathbf{y}, t_* \sim \mathcal{N}(\mathbf{\mu}_*, \mathbf{\Sigma}_*),
\]
\vspace{-1pt}
where:
\vspace{-5pt}
\[
\mathbf{\mu}_* = k(t_*, \mathbf{t}) [k(\mathbf{t}, \mathbf{t}) + \sigma_n^2 I]^{-1} \mathbf{y},
\]
\vspace{-10pt}
is the mean of the predictive distribution and
\[
\mathbf{\Sigma}_* = k(t_*, t_*) - k(t_*, \mathbf{t}) [k(\mathbf{t}, \mathbf{t}) + \sigma_n^2 I]^{-1} k(\mathbf{t}, t_*)
\]
is the variance of the predictive distribution. Here, \( \mathbf{t} \) is the vector of input timestamps, \( \mathbf{y} \) is the vector of observed 3D positions, and \( \sigma_n^2 \) is the noise variance.

In the Gaussian process, the choice of the kernel function \( k(t, t') \) and its parameters significantly affect the smoothness and accuracy of the trajectory estimation. We use the Radial Basis Function (RBF) kernel for its smoothness properties and its ability to model local similarities effectively:

\[
k(t, t') = \sigma_f^2 \exp\left(-\frac{\|t - t'\|^2}{2l^2}\right)
\]

In this kernel:

\begin{itemize}
    \item \( \sigma_f^2 \) controls the vertical variation, or the amplitude, of the function.
    \item \( l \), the length scale parameter, controls the smoothness of the function. A smaller \( l \) results in a function that can vary more rapidly, while a larger \( l \) produces a smoother and more stable function.
\end{itemize}


Through experimentation, we found that a length scale \( l \) of approximately 300, \(\sigma_n^2\) of 2.0, and \(\sigma_f^2\) of 1.0 provided the best results in our UAV trajectory estimation tasks. This value was optimal in capturing the essential patterns of the motion of the UAV while filtering out high-frequency noise. By employing Gaussian Process smoothing with an optimized RBF kernel, we ensure that the trajectory is smoothed.


\begin{figure}[t]
\centering
\includegraphics[width=0.38\textwidth]{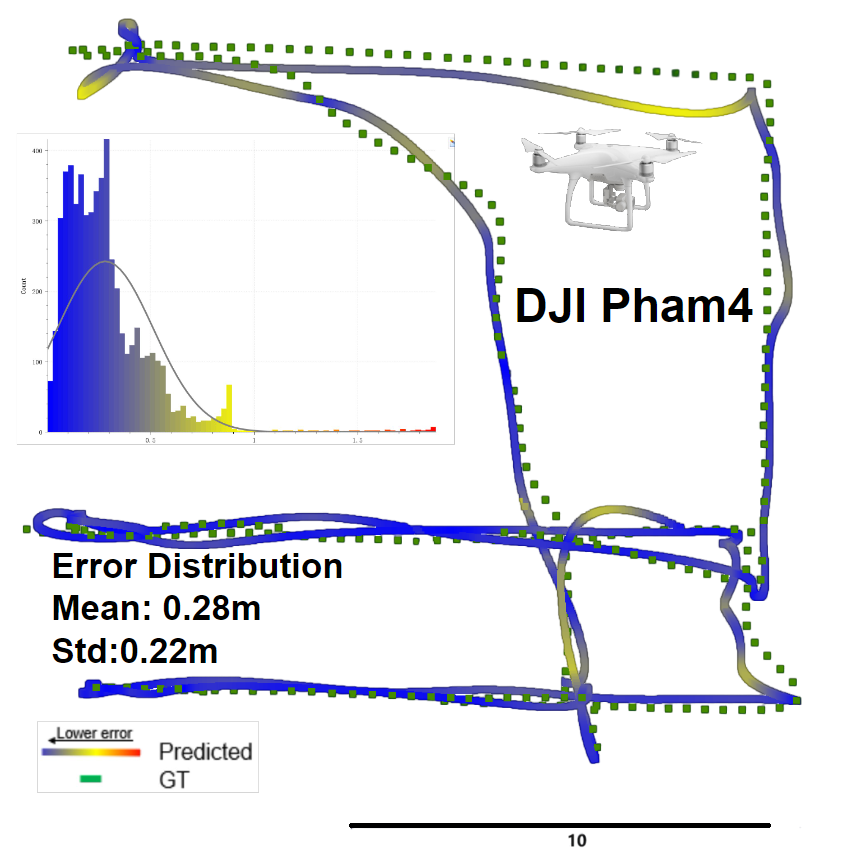}
\vspace{-1em}
\caption{Pham4 Trajectory and Analysis}
\label{fig:pham4_result}
\vspace{-2em}
\end{figure}

\vspace{-5pt}
\subsection{Trajectory Estimation Head}
\vspace{-5pt}
 We estimate the 3D trajectory of UAVs directly without relying on detection or predefined anchor boxes. Instead, we employ pseudo labels derived from high-precision LiDAR trajectory estimates to supervise the training process. Consequently, the training loss for the trajectory estimation head is formulated as follows: 
\vspace{-5pt}
\[
\text{Loss} = \frac{(1-\alpha)}{N} \sum_{i=1}^{N} \left( {\text{P}}^{(i)} - {\xi}^{(i)} \right)^2 + \frac{\alpha}{N} \sum_{i=1}^{N} \left( {\text{P}}^{(i)} - {\Xi}^{(i)} \right)^2
\]
\vspace{-5pt}

where
${P}, {\xi} \text{ and } {\Xi} \in \mathbb{R}^{T \times 3}$. P is the predicted trajectory, $\xi$ is the ground truth trajectory, and $\Xi$ is pseudo labels. If \(\alpha\) is set to 1, our framework is trained in a fully self-supervised mode. If ground truth data is available, setting \(\alpha\) to 0 transitions the system to a supervised mode. It is designed to handle real-world scenarios, adapting to both available ground truth and the need for automatic pseudo label generation.

\begin{table*}[t]
\footnotesize
\renewcommand*{\arraystretch}{1.1}
\centering
\caption{The comparison of performance. Lower \(D_{x,y,z}\) and \(E\) represent better trajectory estimation in m.}
\vspace{-5pt}
\resizebox{0.95\textwidth}{!}{
\begin{tabular}{lccccccccccc}
\hline
\toprule
\textbf{Methods} & \textbf{Category} & \multicolumn{4}{c}{\textbf{Ideal Light Condition}} & \multicolumn{4}{c}{\textbf{Dark Environment}} & \textbf{Mean} & \textbf{GLOPs}\\
\cmidrule(lr){3-6} \cmidrule(lr){7-10}
 &  & $D_x$ & $D_y$ & $D_z$ & $E$ & $D_x$ & $D_y$ & $D_z$ & $E$ & $\bar E$  & \\
\midrule
\textbf{VisualNet}\cite{9814870} & Supervised & 0.24 & 0.39 & 0.32 & 0.65 & 1.98 & 6.10 & 8.13 & 11.45 & 5.05 &  5.67\\
\textbf{DarkNet}\cite{bochkovskiy2020yolov4optimalspeedaccuracy} & Supervised & 0.23 & 0.46 & \textbf{0.23} & 0.63 & 1.84 & 5.50 & 4.57 & 8.31 & 3.98 & 9.31\\
\hline
\textbf{AudioNet}\cite{yang2023av} & Supervised & 0.60 & 1.76 & 1.59 & 2.80 & 0.60 & 1.76 & 1.59 & 2.80 & 2.80 & \textbf{0.005}\\
\textbf{VorasNet}\cite{vora2023dronechase} & Supervised & 0.54 & 1.59 & 1.51 & 2.64 & 0.54 & 1.59 & 1.51 & 2.64 & 2.64 & \textbf{0.005}\\
\textbf{ASDNet}\cite{tao2021someone} & Supervised & 0.31 & 0.69 & 0.44 & 0.99 & 1.13 & 3.39 & 3.92 & 5.82 & 3.41 & 15.30\\
\hline
\textbf{AV-PED}\cite{yang2023av} & Self-Supervised & 0.31 & 0.50 & 0.59 & 0.97 & 0.58 & 1.54 & 2.26 & 3.13 & 2.05 & 5.67\\
\textbf{AV-FDTI}\cite{YANG2024} & Supervised & \textbf{0.13} & \underline{0.31} & 0.38 & \underline{0.58} & \underline{0.35} & \underline{1.06} & \underline{1.10} & \underline{1.89} & \underline{1.24} &  11.37\\
\hline
\textbf{Ours} & Self-Supervised & \underline{0.14} & \textbf{0.26} & \underline{0.25} & \textbf{0.48} & \textbf{0.14} & \textbf{0.26} & \textbf{0.25} & \textbf{0.48} & \textbf{0.48} & \underline{0.0065}\\
\bottomrule
\end{tabular}
}
\label{tab:comparison}
\\
\footnotesize{Best results are highlighted in \textbf{bold}, and second best in \underline{underline}.}
\vspace{-2em}
\end{table*}

\section{Experiments and Evaluation}
\vspace{-5pt}
\subsection{Dataset and Implement Details}

 Our model is trained with a multi-modal anti-UAV dataset\cite{yuan2024MMAUD}. The dataset includes five kinds of drones: Avata, M300, Mavic2, Mavic3, and Pham4. All drones fly in and out of a \(10\,m \times 30\,m \times 25\,m\) region to simulate the drone crossing the border.   

To train our framework, we use Adam as our optimizer with a batch size of 32 and a learning rate of \(1 \times e^{-4}\)  for 100 epochs. All the experiments are conducted on a GeForce GTX 3090 GPU. 
\vspace{-5pt}
\subsection{Evaluation metrics}
\vspace{-5pt}
 In our experiments, we use center distance \(D_x\), \(D_y\), \(D_z\) as our evaluation metrics, computed by the L1 Loss between Predicted trajectory \(P \) and ground truth \(\Xi\) on X, Y,and Z coordinates respectively.

 To better understand the overall network performance, we use \(E\), which stands for Average Position Error(APE) to evaluate the trajectory estimation performance calculated by the root mean square error(RMSE). 

 To test the computational efficiency of our model, the inference time is also calculated to represent the time cost of inferring one audio clip.
\vspace{-5pt}
\subsection{Trajectory Estimation Performance}
\vspace{-5pt}


 The comparison of the estimated trajectory by the proposed method and the ground truth trajectory is shown in Figure \ref{fig:result}. For all UAVs, the green dashed line represents the ground truth trajectory. The gradient color line in the figure represents the estimated trajectory based on audio information. The distribution of the error between the estimated trajectory and the ground truth gives a clear understanding of how well our proposed system is able to handle the task, with a mean of 0.28m and a standard deviation of 0.22m for the Phamtom 4 drone, as shown in Fig. \ref{fig:pham4_result}. It is evident that the estimated trajectory closely aligns with the ground truth in most instances, except for some sharp turns, where slight distortions are observed. This suggests that our solution performs well in estimating the 3D trajectories of UAVs, though there are minor challenges during highly dynamic maneuvers.
\vspace{-5pt}
\subsection{Benchmark Comparisons}
\vspace{-5pt}
 From Table \ref{tab:comparison}, it is clear that our method exhibits a significantly lower average position error compared to all other methods, indicating a superior performance in trajectory prediction. Our methods are compared with two audio-based networks \textbf{AudioNet} and \textbf{Vora's Net}, two visual-based networks \textbf{VisualNet} and \textbf{DarkNet}, and three audio-visual-based networks \textbf{ASDNet}, \textbf{AVped}, and \textbf{AV-FDTI}.

Under ideal light conditions, our proposed method achieves the lowest overall error of 0.48, outperforming other methods in terms of precision. In the dark environment, since our model relies on audio information, its performance is not affected by the light conditions. As a result, the overall error remains the same. Furthermore, under both conditions, our model demonstrates significantly lower error compared to other methods. 
\vspace{-5pt}
\section{Conclusion}
\vspace{-5pt}
 In this paper, we introduce a self-supervised audio anti-UAV framework to track and estimate the trajectory of UAVs. In addition, we developed an unsupervised method to estimate trajectories with highly accurate LiDARs. The unsupervised method served as a Teacher Network, providing pseudo labels to supervise the training of the Audio Perception Network, which acted as the Student Network. Our method provides a new benchmark against all SOTA methods with the highest accuracy without ground truth.

{\footnotesize
\bibliographystyle{IEEEbib}
\bibliography{mybib}
}

\end{document}